%% file: main.tex
\pdfoutput=1

\documentclass{article}

\usepackage[preprint]{neurips_2025}

\usepackage[utf8]{inputenc} 
\usepackage[T1]{fontenc}    
\usepackage{hyperref}       
\usepackage{url}            
\usepackage{booktabs}       
\usepackage{amsfonts}       
\usepackage{nicefrac}       
\usepackage{microtype}      
\usepackage{xcolor}         

\usepackage[textsize=small]{todonotes}

\usepackage{graphicx}
\usepackage{subcaption}

\usepackage{amsmath,amssymb,amsthm}
\usepackage{bbm}

\input{macros}

\newtheorem{theorem}{Theorem}[section]
\newtheorem{lemma}[theorem]{Lemma}

\newtheorem{definition}{Definition}[section]

\title{Generalizability vs. Counterfactual Explainability Trade-Off}

\author{%
    Fabiano Veglianti \\
  Sapienza University of Rome, Italy\\
  \texttt{fabiano.veglianti@uniroma1.it} \\
  \And
  Flavio Giorgi \\
  Sapienza University of Rome, Italy \\
  \texttt{giorgi@di.uniroma1.it} \\
  \And
  Fabrizio Silvestri \\
  Sapienza University of Rome, Italy \\
  \texttt{fsilvestri@diag.uniroma1.it} \\
  \And
  Gabriele Tolomei \\
  Sapienza University of Rome, Italy \\
  \texttt{tolomei@di.uniroma.it} \\
}

\begin{document}

\maketitle

\begin{abstract}
In this work, we investigate the relationship between model generalization and counterfactual explainability in supervised learning. We introduce the notion of $\varepsilon$-\textit{valid counterfactual probability} ($\varepsilon$-VCP) -- the probability of finding perturbations of a data point within its $\varepsilon$-neighborhood that result in a label change. We provide a theoretical analysis of $\varepsilon$-VCP in relation to the geometry of the model's decision boundary, showing that $\varepsilon$-VCP tends to increase with model overfitting. Our findings establish a rigorous connection between poor generalization and the ease of counterfactual generation, revealing an inherent trade-off between generalization and counterfactual explainability. 
Empirical results validate our theory, suggesting $\varepsilon$-VCP as a practical proxy for quantitatively characterizing overfitting.
\end{abstract}

\input{intro}
\input{related}
\input{background}
\input{theory}

\input{experiments}

\input{limitations}
\input{conclusion}

\bibliographystyle{plainnat}
\bibliography{references}
\newpage
\appendix

\input{supplementary}


\end{document}

%% file: macros.tex
\newcommand{\Prob}{\mathbb{P}}
\newcommand{\E}{\mathbb{E}}
\newcommand{\R}{\mathbb{R}}
\newcommand{\X}{\mathcal{X}}
\newcommand{\Y}{\mathcal{Y}}
\newcommand{\ball}{\mathcal{B}}
\newcommand{\paramspace}{\boldsymbol{\Theta}}
\newcommand{\params}{\boldsymbol{\theta}}

\newcommand{\model}{f_{\params}}
\newcommand{\hyp}{h_{\params}}

\newcommand{\inst}{\boldsymbol{x}}
\newcommand{\cfinst}{\widetilde{\boldsymbol{x}}}

\newcommand{\dataset}{\mathcal{D}}

\newcommand{\argmin}{\text{arg\,min}}

%% file: intro.tex
\section{Introduction}
\label{sec:intro}
One of the key challenges in machine learning is developing models that can generalize their predictions to new, unseen data beyond the scope of the training set. 
When the predictive accuracy of a model on the training set far exceeds that on the test set, this indicates a phenomenon known as \textit{overfitting}.
In general, the impact of overfitting is more pronounced for highly complex models like recent deep neural networks with billions of parameters. 
To compensate for the risk of overfitting, these models require massive amounts of training data, which may not always be feasible. 

While several techniques -- such as \textit{early stopping}, \textit{data augmentation}, and \textit{regularization} -- have been developed to mitigate overfitting, a quantitative characterization of the phenomenon is still lacking. As a result, overfitting is typically identified only empirically, by observing discrepancies between training and test errors.

In this work, we offer an entirely new perspective on model overfitting, establishing a connection with the ability to generate \textit{counterfactual examples} \citep{wachter2017hjlt}. The notion of counterfactual examples has been successfully used, for instance, to attach post-hoc explanations for predictions of individual instances in the form: ``\textit{If A had been different, B would \textbf{not} have occurred}'' \citep{stepin2021survey}. 
Generally, finding the counterfactual example for an instance resorts to searching for the (minimal) perturbation of the input that crosses the decision boundary induced by a trained model. This task often reduces to solving a constrained optimization problem.

\begin{figure}
\centering
\begin{subfigure}[htb!]{.2\linewidth}
\includegraphics[width=\linewidth]{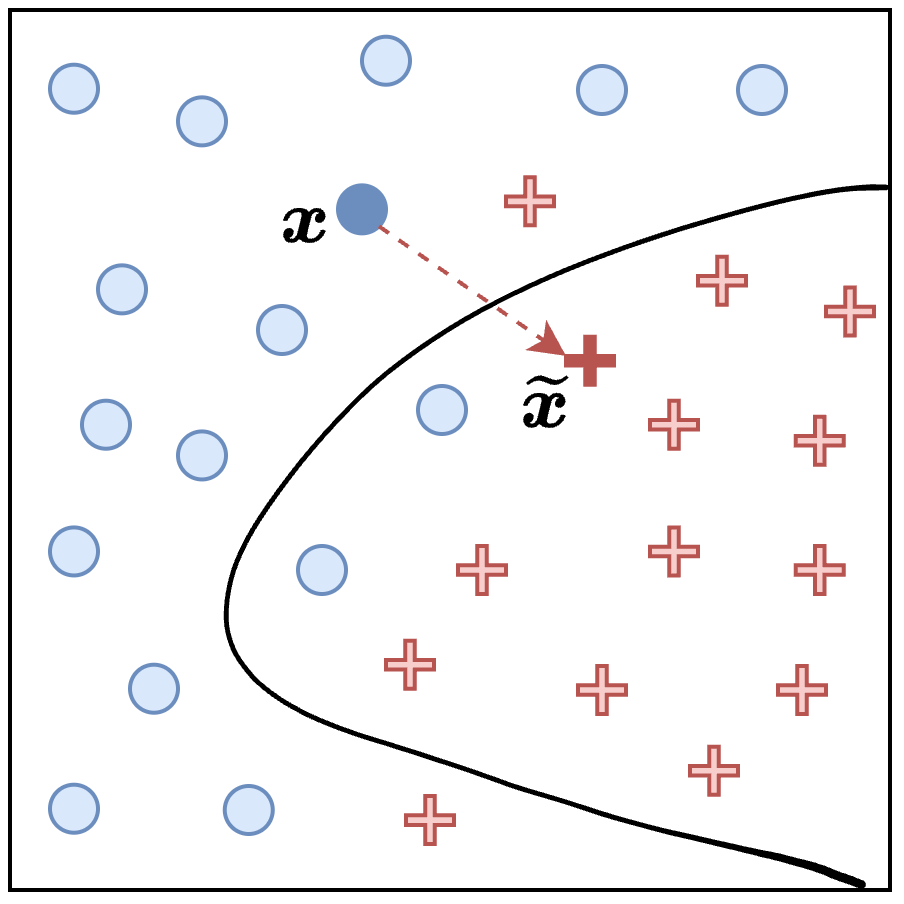}
\caption{No overfitting}
\end{subfigure}
\qquad
\begin{subfigure}[h]{.2\linewidth}
\includegraphics[width=\linewidth]{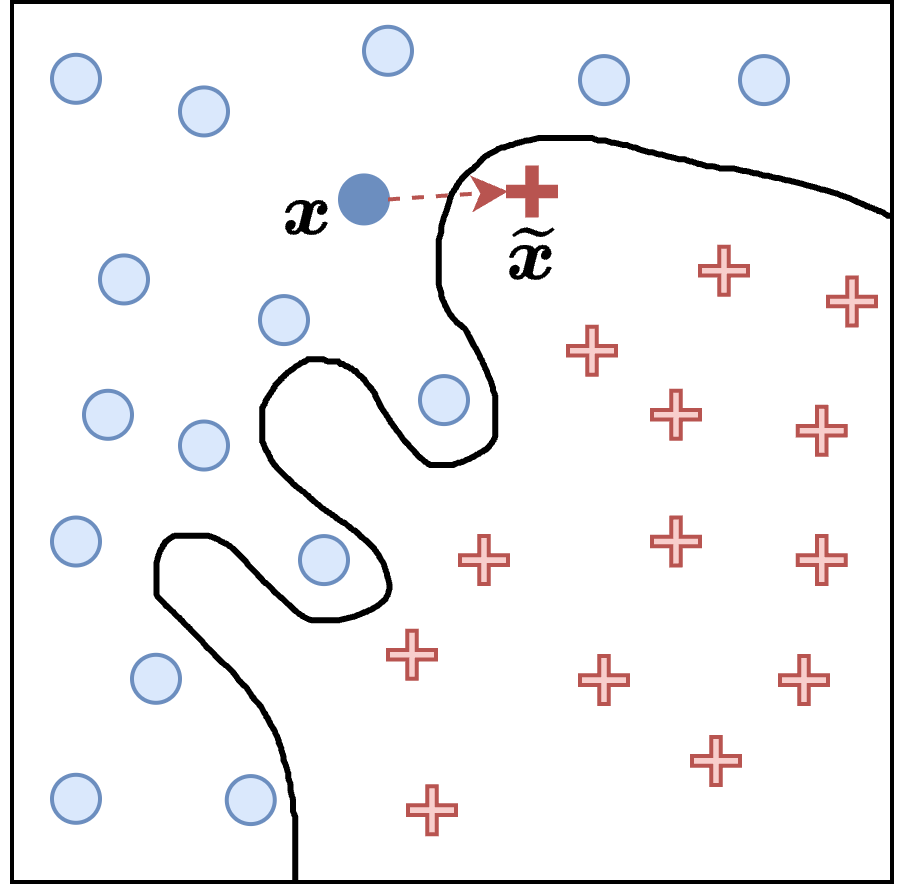}
\caption{Overfitting}
\end{subfigure}%
\caption{Distance between an input data point ($\inst$) and its counterfactual example ($\cfinst$): On average, this may be higher for a well-trained model (a) than an overfitted model (b).}
\label{fig:intuition}
\end{figure}

In the presence of a strong degree of model overfitting, the decision boundary learned becomes a highly convoluted surface, up to the point where it perfectly separates every training input sample. 
Therefore, each data point, on average, is ``closer'' to the decision boundary, making it easier to find the best counterfactual example. 
The intuitive explanation of this claim is illustrated in Figure~\ref{fig:intuition}.

Following this idea, we theoretically analyze the connection between model's generalizability and the ability of finding counterfactual examples.
The key novel contributions of our work are as follows:
\begin{itemize}
\item[$(i)$] We are the first to investigate the relationship between model generalizability and counterfactual explanations.
\item[$(ii)$] We introduce the concept of $\varepsilon$-valid counterfactual probability ($\varepsilon$-VCP).
\item[$(iii)$] We establish theoretical results linking $\varepsilon$-VCP to the model's decision boundary, revealing a trade-off between generalization and counterfactual explainability.
\item[$(iv)$] We propose the average $\varepsilon$-VCP as a \textit{new} proxy to quantify model generalizability, demonstrating that it increases with overfitting.
\item[$(v)$] We validate our findings empirically. The source code for our experiments is available at: \url{https://anonymous.4open.science/r/generalizability_ce_tradeoff-631F/README.md}.
\end{itemize}

The remainder of this paper is structured as follows. Section~\ref{sec:related} summarizes related work. Section~\ref{sec:background} reviews background and preliminary concepts. In Section~\ref{sec:theory}, we describe our theoretical analysis that is validated empirically in Section~\ref{sec:experiments}. 
We discuss the limitations of our work along with possible future directions in Section~\ref{sec:limitations}.
Finally, we conclude in Section~\ref{sec:conclusion}.

%% file: related.tex
\section{Related Work}
\label{sec:related}

\noindent \textit{\textbf{Model Overfitting and Margin Theory.}}
%
%
Various strategies have been proposed in the literature to reduce the risk of overfitting, such as \textit{early stopping} \citep{raskutti2011ccc,caruana2000neurips}, \textit{data augmentation} \citep{sun2014cvpr,karystinos2000tnn,yip2008bio}, and \textit{regularization} \citep{warde-farley2014iclr,lin2023neurips}.
When a model overfits, it tightly conforms to training data, including noise or rare patterns. This leads to sharper decision boundaries near training points and smaller distances between data points and the decision boundary.
This is well supported by studies on margin theory, showing that overfitting results in reduced margins around training samples on average (e.g., see \cite{mason1998neurips,neyshabur2017neurips,wu2019arxiv,shamir2021dimpled}).
\noindent \textit{\textbf{Counterfactual Examples.}} Counterfactual examples are the cornerstone that offer valuable explanations into model predictions by providing alternative scenarios under which the prediction would change \citep{wachter2017hjlt}. 
Several studies have explored the use of counterfactual examples to enhance the interpretability of complex machine learning models, such as ensembles of decision trees \citep{tolomei2017kdd,tolomei2021tkde,lucic2022aaai} and deep neural networks (DNNs) \citep{le2020kdd}, including graph neural networks (GNNs) \citep{lucic2022aistats}. Counterfactual instances may elucidate the underlying decision-making process of black-box models, enhancing trust and transparency \citep{guidotti2018arxiv,karimi2020pmlr,chen2022cikm}.
\\
%
Counterfactual examples have also been leveraged to enhance the robustness of machine learning models against adversarial attacks \citep{brown2018arxiv}. Indeed, there is a strict relationship between adversarial and counterfactual examples \citep{freiesleben2022mm}, although their primary goals are divergent. While both adversarial and counterfactual examples perturb input data to influence model predictions, adversarial examples are crafted to jeopardize the model, whereas counterfactual examples are generated to understand the model's behavior.
By generating instances that are semantically similar to the original input but induce different model predictions, \citet{he2019acsac} aim to improve model robustness against adversarial perturbations. These counterfactual examples serve as natural adversaries, enabling the model to learn more robust decision boundaries against malicious attacks.
%
%
\\
Recently, generative models have been used to build realistic counterfactual instances for data augmentation and model improvement \citep{ganin2016jmlr,hjelm2019iclr}. 

To the best of our knowledge, this is the first study to link overfitting and margin theory with counterfactual examples, thus quantifying model generalizability through counterfactual explainability.

%% file: background.tex
\section{Background and Preliminaries}
\label{sec:background}
Let $\model: \X \rightarrow \Y$ denote a predictive model parameterized by learnable weights $\params \in \paramspace$. Given an input $\inst \in \X$, the model outputs a prediction $y \in \Y$.

In the following, we focus on binary classification, although our analysis can generalize to multiclass classification and regression.
Specifically, we let $\X \subseteq \R^n$ and $\Y = \{0, 1\}$ and consider a real-valued scoring function $\hyp: \X \rightarrow \R$ (e.g., pre-activation logits). The classifier's decision rule is threshold-based: $\model(\inst) = \mathbbm{1}\{\inst \in \X : \hyp(\inst) \geq 0\}$, where $\mathbbm{1}$ is the indicator function. 
\begin{definition}[Decision Boundary]
Let $\model(\inst)$ be a classifier as introduced above. The decision boundary induced by $\hyp$ is defined as follows:
\begin{equation}
\label{eq:decision-boundary}
\mathcal{H}_{\params} = \{\inst \in \X: \hyp(\inst) = 0\}.
\end{equation}
\end{definition}
\begin{definition}[Geometric Margin]
\label{def:geo-margin}
Let $\model$ be a classifier and $\inst_i \in \X$ a generic input. The geometric margin of $\inst_i$, denoted as $\gamma_i$, is the distance from $\inst_i$ to the decision boundary induced by $\hyp$:
\begin{equation}
\label{eq:geo-margin}
\gamma_i = \inf\{||\inst_i - \inst'||:\inst' \in \mathcal{H}_{\params}\} = \inf\{||\inst_i - \inst'||:\inst' \in \X, \hyp(\inst') = 0\}.
\end{equation}
\end{definition}
\begin{definition}[Counterfactual Example]
\label{def:cf}
Let $\model$ be a classifier and $\inst_i \in \X$ a generic input. We consider a counterfactual example for $\inst_i$ as a data point $\cfinst_i$ obtained from an additive perturbation $\boldsymbol{r} \in \X$ -- i.e., $\cfinst_i = \inst_i + \boldsymbol{r}$ -- such that $\model(\cfinst_i) \neq \model(\inst_i)$. The optimal counterfactual example for $\inst_i$ is the minimal perturbed counterfactual example $\cfinst_i^* = \inst_i + \boldsymbol{r}^*$, namely:
\begin{equation}
\begin{aligned}
\label{eq:cf-example}
\cfinst_i^* = \inst_i + \boldsymbol{r}^*, \text{ where } \boldsymbol{r}^* = \underset{\boldsymbol{r} \in \X}\argmin  &~\delta(\inst_i + \boldsymbol{r}, \inst_i), \
\text{ s.t.: }\model(\inst_i + \boldsymbol{r}) \neq \model(\inst).
\end{aligned}
\end{equation}
\end{definition}
Since the concept of ``minimality'' can vary depending on factors such as the application domain, we adopt a general approach by defining $\delta:\X\times \X \mapsto \R_{\geq 0}$ as a function that captures some notion of distance between the original input and its counterfactual. In practice, however, $\delta$ is frequently instantiated as the $L^2$-norm of the displacement vector between the two, i.e., $||\cfinst_i - \inst_i|| = ||\boldsymbol{r}||$.

While multiple counterfactuals may exist for a single input, we consider any $\cfinst$ such that $\model(\cfinst) \neq \model(\inst)$ as \textit{valid}. Notions of plausibility -- i.e., realism of generated counterfactual examples -- are critical but orthogonal to our investigation. We focus purely on validity and its relation to generalization.

\begin{definition}[$\varepsilon$-Valid Counterfactual Example -- $\varepsilon$-VCE]
\label{def:evcf}
Let $\model$ be a classifier, $\inst_i \in \X$ a generic input, and $\varepsilon \in \R_{>0}$ a fixed threshold. Any $\cfinst_i = \inst_i + \boldsymbol{r}$ is an $\varepsilon$-valid counterfactual example for $\inst_i$ if: $(i)$ $\model(\cfinst_i) \neq \model(\inst_i)$ and  $(ii)$ $||\boldsymbol{r}|| \leq \varepsilon$.
\end{definition}
From Def. \ref{def:geo-margin} and \ref{def:evcf}, we can derive the region where the set of $\varepsilon$-valid counterfactual examples for a given input $\inst_i$ might exist, providing $\varepsilon \geq \gamma_i$.  More formally:
\begin{lemma}[$\varepsilon$-Valid Counterfactual Shell]
Let $\model$ be a classifier, $\inst_i \in \X$ a generic input, $\gamma_i$ the geometric margin from $\inst_i$ to the decision boundary induced by $\hyp$, and $\varepsilon \in \R_{>0}$ a fixed threshold, such that $\varepsilon \geq \gamma_i$. 
The only possible region where $\varepsilon$-valid counterfactuals for $\inst_i$ can be found is:
\begin{equation}
\label{eq:shell}
\mathcal{S}(\inst_i, \gamma_i, \varepsilon) = \ball(\inst_i, \varepsilon) \setminus \ball(\inst_i, \gamma_i) = \{\inst \in \X : \gamma_i \leq ||\inst - \inst_i|| < \varepsilon\},
\end{equation}
where $\ball(\inst_i, \rho) = \{\inst \in \X: ||\inst - \inst_i|| < \rho\}$ represents the open $n$-ball restricted on $\X$ and centered at $\inst_i$ with radius $\rho$. We refer to the region defined in \eqref{eq:shell} as the $\varepsilon$-valid counterfactual shell.
\begin{proof}
Every data point in the open $n$-ball $\ball(\inst_i, \gamma_i)$ must have the same label as $\inst_i$, i.e., $\model(\inst_i)$, as they do not cross the decision boundary $\mathcal{H}_{\params}$ by definition. Indeed:
\[
\forall \boldsymbol{r} \in \X \text{ s.t. } ||\boldsymbol{r}|| < \gamma_i \Rightarrow \model(\inst_i + \boldsymbol{r}) = \model(\inst_i).
\]
Hence, any $\varepsilon$-valid counterfactual $\cfinst_i$ for $\inst_i$ must lie outside the ball $\ball(\inst_i, \gamma_i)$ -- that is, $||\cfinst_i - \inst_i|| \geq \gamma_i$ -- but still within the ball $\ball(\inst_i, \varepsilon)$ -- i.e., $||\cfinst_i - \inst_i|| < \varepsilon$. In other words, $\cfinst_i$ must belong to the counterfactual shell $\mathcal{S}(\inst_i, \gamma_i, \varepsilon)$.
Note that if $\varepsilon < \gamma_i$, this shell is trivially empty.
\end{proof}
\end{lemma}

%% file: theory.tex
\section{The Impact of Counterfactual Examples on Model Generalizability}
\label{sec:theory}

\subsection{Intuition}
Consider a dataset $\dataset = \{(\inst_i, y_i)\}_{i=1}^m$ used to train a sequence of models $\{f_{\params_t}\}_{t=0}^T$, where each model $f_{\params_t}$ corresponds to the model parameters at training epoch $t$ and achieves a training accuracy of $\alpha_t$. We assume that accuracy increases over time, i.e., $\alpha_0 < \alpha_1 < \ldots < \alpha_T$. In practice, such a sequence may consist of intermediate checkpoint models saved at different stages of training.

We hypothesize that the average fraction of training points for which we can find $\varepsilon$-valid counterfactuals increases with model training accuracy. That is, for two models $f_{\params_{t'}}$ and $f_{\params_t}$ with $\alpha_{t'} > \alpha_t$, we expect that:
\begin{equation}
\label{eq:claim}
\mathbb{E}[|\{\inst_i: \exists~\varepsilon\text{-VCE under }f_{\params_{t'}}\}|/m] > \mathbb{E}[|\{\inst_i: \exists~\varepsilon\text{-VCE under }f_{\params_{t}}\}|/m].
\end{equation}
Highly accurate models exhibit more complex decision boundaries that fit the training data more tightly. Indeed, \citet{wu2019arxiv} proved that the average distance from a data point to the decision boundary (i.e., margin) decreases as training progresses. Hence, the likelihood of finding nearby counterfactuals increases.

In essence, a model for which finding counterfactual examples is ``too easy'' may indicate a risk of overfitting. Thus, a trade-off must exist between the model's generalizability and its counterfactual explainability, which we aim to investigate further in this study.

\subsection{$\varepsilon$-Valid Counterfactual Probability}
\label{subsec:epsilon-VCF}
To verify our claim, we need to characterize better what we mean by the ease of finding an $\varepsilon$-valid counterfactual example for a given data point and, therefore, for a full training set of data points.

Let us consider the generic predictive model $\model$ trained on $\dataset = \{(\inst_i, y_i)\}_{i=1}^m$.
Suppose we associate with each training data point $\inst_i$ a binary random variable $X_i\in \{0,1\}$, which indicates whether there exists an $\varepsilon$-valid counterfactual example $\cfinst_i$ for $\inst_i$. 
Therefore, each $X_i$ follows a Bernoulli distribution, i.e., $X_i \sim \text{Bernoulli}(p_i^{\varepsilon})$, whose probability mass function is defined as below.
\begin{equation}
\Prob(X_i = k) = p_{X_i}(k;p_i^{\varepsilon}) = \begin{cases} p_i^{\varepsilon} &\text{, if }k=1
\\ 
1-p_i^{\varepsilon} &\text{, if } k=0.
\end{cases}
\end{equation}
We refer to $p_i^{\varepsilon}$ as the (sample-level) $\varepsilon$-\textit{valid counterfactual probability} ($\varepsilon$-VCP) for $\inst_i$.
\begin{definition}[$\varepsilon$-Valid Counterfactual Probability -- $\varepsilon$-VCP]
Let $\model$ be a trained classifier, $\inst_i \in \X$ a generic input sample, and $\varepsilon \in \R_{>0}$ a fixed threshold. 
The $\varepsilon$-\textit{valid counterfactual probability} ($\varepsilon$-VCP) for $\inst_i$ is the probability that a random perturbation $\boldsymbol{r}$ drawn from a probability distribution $\varphi$ supported on the counterfactual shell $\mathcal{S}(\inst_i, \gamma_i, \varepsilon)$ defined in \eqref{eq:shell} and applied to $\inst_i$ -- i.e.,  $\cfinst_i = \inst_i + \boldsymbol{r}$ -- leads to an $\varepsilon$-valid counterfactual example for $\inst_i$. We denote this probability by $p_i^{\varepsilon}$:
\begin{equation}
\label{eq:evcp}
   p_i^{\varepsilon} = \Prob_{\boldsymbol{r} \sim \varphi}[\model(\inst_i + \boldsymbol{r})\neq \model(\inst_i)] = \int_{\mathcal{S}(\inst_i,\gamma_i,\varepsilon)}
    \mathbbm{1}\bigl\{\model(\inst_i + \boldsymbol{r})\neq \model(\inst_i)\bigr\}\,
    q_{\varphi}(\boldsymbol{r})\,\mathrm{d}\boldsymbol{r},
\end{equation}
where $q_{\varphi}$ is the probability density function associated with $\varphi$.
\end{definition}
\textbf{\textit{Remark:}} This formulation is flexible, as it makes no assumptions about the underlying distribution $\varphi$ used to sample legitimate perturbations $\boldsymbol{r}$. Specific settings will be explored in the following sections.

Furthermore, let $\bar{X}$ denote the fraction of training points for which an $\varepsilon$-valid counterfactual example exists. Formally, $\bar{X}=\frac{1}{m}\sum_{i=1}^m X_i$ is the average of $m$ Bernoulli random variables. By the linearity of expectation, we can calculate:
\[
\E[\bar{X}] = \E\Big[ \frac{1}{m}\sum_{i=1}^m X_i\Big] = \frac{1}{m}\sum_{i=1}^m \E[X_i],
\]
where $\E[X_i] = p_i^{\varepsilon}$ and, therefore, $\E[\bar{X}] = \bar{p}^{\varepsilon}$ is the \textit{average} $\varepsilon$-VCP across the entire dataset.
\\
Note that the random variable $Z=\sum_{i=1}^m X_i$ follows a Poisson-Binomial distribution, since $X_i$'s are independent but not necessarily identically distributed. Consequently, $\bar{X} = \frac{Z}{m}$ is also a Poisson-Binomial random variable, scaled by a factor of $1/m$.

We can formalize the intuitive claim defined in \eqref{eq:claim} based on the average $\varepsilon$-VCP as follows:
\begin{equation}
\label{eq:claim2}
\E[\bar{X}_{\params_{t'}}] > \E[\bar{X}_{\params_t}],
\end{equation}
where $\bar{X}_{\params_{t'}}$ and $\bar{X}_{\params_{t}}$ are the average $\varepsilon$-VCP computed across $m$ samples based on the trained models $f_{\params_{t'}}$ and $f_{\params_t}$, respectively, such that $\alpha_{t'} > \alpha_t$ (i.e., the model $f_{\params_{t'}}$ achieves higher accuracy).

To validate \eqref{eq:claim2}, we need to estimate each $p_i^{\varepsilon}$ and, therefore, the average $\varepsilon$-VCP $\bar{p}^{\varepsilon}$.

\subsection{Linking $\varepsilon$-Valid Counterfactual Probability to the Decision Boundary}
\label{subsec:margin2vcp}
In this section, we explore the relationship between the probability of generating $\varepsilon$-valid counterfactuals and the geometry of the model's decision boundary. 

We present theoretical results concerning $p_i^{\varepsilon}$, each incorporating progressively more detailed geometric insights about the model. In particular, we distinguish between two settings: models with a \textit{linear} decision boundary and those with a \textit{non-linear} one.

\subsubsection{Case 1: Linear Models}
We start our discussion from the simplest setting represented by a linear classifier $\model$, whose scoring function is linear and defined as $\hyp(\inst) = \params^T \inst$.
\begin{theorem}
\label{thm: epsilon-vcp linear model}
[$\varepsilon$-VCP for Linear Models Under Generic Perturbation Distribution $\varphi$]
Let $\model$ be a linear classifier and $\inst_i \in \X$ a generic input. The decision boundary $\mathcal{H}_{\params}$ induced by $\hyp$ is the hyperplane tangent to $\bar{\inst}_i$, which is the closest data point to $\inst_i$ on the boundary, i.e., $\bar{\inst}_i \in \mathcal{H}_{\params}$. Since $\varphi$ assigns non-zero density only to points in the counterfactual shell $\mathcal{S}(\inst_i; \gamma_i, \varepsilon)$, i.e., $\text{supp}(\varphi) \subseteq \mathcal{S}(\inst_i; \gamma_i, \varepsilon)$, then:
\[
p_i^{\varepsilon}
= \displaystyle
  \int_{\mathcal{C}(\inst_i,\gamma_i,\varepsilon)}
    q_{\varphi}(\boldsymbol{r})\,\mathrm{d}\boldsymbol{r}, 
\]
where $\mathcal{C}(\inst_i, \gamma_i, \varepsilon)$ is the spherical cap on the open $n$-ball $\ball(\inst_i, \varepsilon)$, centered around $\inst_i$ with angular radius $\arccos(\gamma_i/\varepsilon)$.
\begin{proof}
Let $\boldsymbol{u}_i = (\bar{\inst}_i - \inst_i)/||\bar{\inst}_i - \inst_i||$ be the unit direction vector pointing from $\inst_i$ to the closest point on the decision boundary $\bar{\inst}_i$. Note that $||\bar{\inst}_i - \inst_i|| = \gamma_i$. Furthermore, consider the generic unit direction vector $\boldsymbol{u} \in \mathbb{S}^{n-1}$, where $\mathbb{S}^{n-1} = \{\boldsymbol{u}\ \in \R^n: ||\boldsymbol{u}|| = 1\}$ is the unit $n$-dimensional sphere. 
In addition, since $\hyp$ is linear, the decision boundary is the hyperplane tangent to $\bar{\inst}_i$, i.e., $\mathcal{H}_{\params} = \{\inst \in \X : \langle \bar{\inst}_i - \inst, \boldsymbol{u}_i\rangle = 0\}$, where $\langle \cdot, \cdot \rangle$ indicates the dot product. The signed distance from the generic perturbation $\inst_i + \boldsymbol{r}$ and the hyperplane $\mathcal{H}_{\params}$ corresponds to $\langle \boldsymbol{r}, \boldsymbol{u}_i \rangle - \gamma_i$, where $\langle \boldsymbol{r}, \boldsymbol{u}_i \rangle$ is the scalar projection of the perturbation along the unit direction vector pointing towards the smallest distance to the decision boundary. So, the perturbed point $\inst_i + \boldsymbol{r}$ indeed crosses the boundary, i.e., it is an $\varepsilon$-valid counterfactual example for $\inst_i$, if and only if $\langle \boldsymbol{r}, \boldsymbol{u}_i \rangle \geq \gamma_i$. 
Therefore:
\[
\langle \boldsymbol{r}, \boldsymbol{u}_i \rangle \geq \gamma_i \Leftrightarrow \langle \rho\boldsymbol{u}, \boldsymbol{u}_i \rangle \geq \gamma_i \Leftrightarrow \rho\langle \boldsymbol{u}, \boldsymbol{u}_i \rangle \geq \gamma_i.
\]
Notice that $\langle \boldsymbol{u}, \boldsymbol{u}_i \rangle = ||\boldsymbol{u}||||\boldsymbol{u}_i||\cos(\beta)$, where $\beta$ is the angle between $\boldsymbol{u}$ and $\boldsymbol{u}_i$. Since both $\boldsymbol{u}$ and $\boldsymbol{u}_i$ are unit-length vectors, $\langle \boldsymbol{u}, \boldsymbol{u}_i \rangle = \cos(\beta)$.
Hence:
\[
\rho \cos(\beta) \geq \gamma_i \Leftrightarrow \cos(\beta) \geq \frac{\gamma_i}{\rho} \geq \frac{\gamma_i}{\varepsilon}. 
\]
The last inequality comes from the fact that $\rho < \varepsilon$.
So, only perturbation directions $\boldsymbol{r}$ forming an angle $\beta$ with $\boldsymbol{u}_i$ such that $\cos(\beta) \geq \frac{\gamma_i}{\varepsilon}$ can lead to $\varepsilon$-valid counterfactuals for $\inst_i$.

This defines a \textit{spherical cap} $\mathcal{C}(\inst_i, \gamma_i, \varepsilon)$ on the open $n$-ball $\ball(\inst_i, \varepsilon)$, 
centered around $\inst_i$ with angular radius $\arccos(\gamma_i/\varepsilon)$:
\[
\mathcal{C}(\inst_i, \gamma_i, \varepsilon) = \left\{ \boldsymbol{r} \in \X \;\middle|\; \gamma_i \leq \|\boldsymbol{r}\| < \varepsilon \;\text{and}\; \langle \boldsymbol{r}, \boldsymbol{u}_i \rangle \geq \gamma_i \right\}.
\]
Hence:
\[
\mathbbm{1}\bigl\{\model(\inst_i + \boldsymbol{r})\neq \model(\inst_i)\bigr\}=
\begin{cases}
1, \text{ if } \boldsymbol{r} \in \mathcal{C}(\inst_i, \gamma_i, \varepsilon)\\
0, \text{ otherwise.}
\end{cases}
\]
Therefore, from \eqref{eq:evcp}, we obtain the following:
\begin{equation}
\label{eq:evcp-linear}
p_i^{\varepsilon}
=
  \displaystyle
  \int_{\mathcal{S}(\inst_i,\gamma_i,\varepsilon)}
    \mathbbm{1}\bigl\{\model(\inst_i + \boldsymbol{r})\neq \model(\inst_i)\bigr\}\,
    q_{\varphi}(\boldsymbol{r})\,\mathrm{d}\boldsymbol{r} = \displaystyle
  \int_{\mathcal{C}(\inst_i,\gamma_i,\varepsilon)}
    q_{\varphi}(\boldsymbol{r})\,\mathrm{d}\boldsymbol{r}.
\end{equation}
\end{proof}
\end{theorem}
In situations where no assumptions can be made about the underlying distribution $\varphi$ used to sample perturbation vectors -- and thus to construct counterfactuals -- it is common practice to assume a uniform distribution. Although this approach can lead to suboptimal or unrealistic counterfactuals, it remains widely used in the absence of prior knowledge about the true data distribution (e.g., \citet{rawal2020beyond}). More sophisticated alternatives do exist, including methods based on generative models such as variational autoencoders \citep{pawelczyk2020learning} and generative adversarial networks \citep{nemirovsky2020countergan}. We discuss the implications of this assumption on $\varphi$ in Section~\ref{sec:limitations}.

\begin{theorem}\label{thm: epsilon-vcp linear model uniform distribution}[$\varepsilon$-VCP for Linear Models Under Uniform Perturbation Distribution $\varphi$] If the random perturbation vectors $\boldsymbol{r}$ are sampled uniformly, i.e., $\varphi = \text{Uniform}(\mathcal{S}(\inst_i; \gamma_i, \varepsilon))$, then:
\begin{equation}
\label{eq:evcp-lin-uniform}
    p_i^{\varepsilon} = \frac{1}{2}  \frac{\varepsilon^n}{\varepsilon^n - \gamma_i^n} I\Big(1-\Big(\frac{\gamma_i}{\varepsilon}\Big)^2;\frac{n+1}{2},\frac{1}{2}\Big),
\end{equation}
where $I(x;a,b)$ is the regularized incomplete Beta function, s.t. $x=1-(\frac{\gamma_i}{\varepsilon})^2$, $a=\frac{n+1}{2}$, and $b=\frac{1}{2}$.
\begin{proof}
Notice that if $q_{\varphi}(\boldsymbol{r})$ is uniform in $\mathcal S(\inst_i,\gamma_i,\varepsilon)$ then $q_{\varphi}(\boldsymbol{r}) = \frac{1}{V(\mathcal S(\inst_i,\gamma_i,\varepsilon))}$, where $V(\cdot)$ denotes volume. Hence, \eqref{eq:evcp-linear} reduces to:
\[
p_i^{\varepsilon}
= \frac{1}{{
  \displaystyle
  V\bigl(\mathcal S(\inst_i,\gamma_i,\varepsilon)\bigr)
}} \displaystyle
  \int_{\mathcal{C}(\inst_i,\gamma_i,\varepsilon)}
    \mathrm{d}\boldsymbol{r} 
    \, =  \frac{
  \displaystyle
  V\bigl(\mathcal{C}(\inst_i,\gamma_i,\varepsilon)\bigr)
}{
  \displaystyle
  V\bigl(\mathcal S(\inst_i,\gamma_i,\varepsilon)\bigr)
}.\,
\]
Since perturbations are drawn uniformly, calculating the $\varepsilon$-VCP reduces to computing the probability that $\boldsymbol{r}$ is sampled from the spherical cap above.
In other words, we need to compute the fraction of the volume of the spherical cap over the total volume of the counterfactual shell, namely:
\[
p^{\varepsilon}_i = \frac{V(\mathcal{C}(\inst_i, \gamma_i, \varepsilon))}{V(\mathcal{S}(\inst_i, \gamma_i, \varepsilon))} = \frac{\frac{1}{2}  C_n \varepsilon^n  I\Big(1-\Big(\frac{\gamma_i}{\varepsilon}\Big)^2;\frac{n+1}{2},\frac{1}{2}\Big)}{C_n (\varepsilon^n - \gamma_i^n)} = \frac{1}{2} \frac{\varepsilon^n}{\varepsilon^n - \gamma_i^n} I\Big(1-\Big(\frac{\gamma_i}{\varepsilon}\Big)^2;\frac{n+1}{2},\frac{1}{2}\Big),
\]
where $C_n = \frac{\pi^{n/2}}{\Gamma(1 + n/2)}$ is the volume of the unit $n$-ball (see \citet{li2010concise}).
\end{proof}
\end{theorem}
\textit{\textbf{Remark \#1:}} The result above captures the intuition that, under our assumptions, $p_i^{\varepsilon} \rightarrow \frac{1}{2}$ as $\gamma_i$ approaches $0$. Formally:
\[
\lim_{\gamma_i \rightarrow 0} p_i^{\varepsilon} = \frac{1}{2}.
\]
Indeed, when $\gamma_i \rightarrow 0$, $\frac{\varepsilon^n}{\varepsilon^n - \gamma_i^n} \rightarrow \frac{\varepsilon^n}{\varepsilon^n} = 1$. Furthermore, $I\Big(1-\Big(\frac{\gamma_i}{\varepsilon}\Big)^2;\frac{n+1}{2},\frac{1}{2}\Big) \rightarrow I\Big(1;\frac{n+1}{2},\frac{1}{2}\Big) = 1$. Overall, $p_i^{\varepsilon} = \frac{1}{2} \cdot 1 \cdot 1 = \frac{1}{2}$. This means that as the point $\inst_i$ gets arbitrarily close to the decision boundary (i.e., $\gamma_i \rightarrow 0$), the chance of randomly sampling a valid counterfactual within the shell becomes $50\%$, which makes sense geometrically: a perfectly centered shell around the decision boundary will intersect it in about half the directions under uniform sampling. 
Conversely, as $\varepsilon \rightarrow \gamma_i$, the value of $p_i^{\varepsilon}$ vanishes to $0$ (please refer to Appendix~\ref{app:theorem-non-linear} for further details).
\\
\textit{\textbf{Remark \#2:}} Notice that when $\model$ is linear, the geometric margin $\gamma_i$ -- the minimum distance between a generic input $\inst_i$ and the hyperplane $\mathcal{H}_{\params}$ induced by $\hyp$ -- can be computed analytically. This is the Euclidean norm of the orthogonal projection of the vector $\inst_i - \inst'$ (for any $\inst' \in \mathcal{H}_{\params}$) onto the normal vector of the hyperplane. 

\subsubsection{Case 2: Non-Linear Models}
\label{subsubsec:non-linear}
We now extend to the more general case where $\model$ is a non-linear classifier, i.e., when the decision boundary induced by $\hyp$ is non-linear. 
A possible way to handle such a case is to consider the decision boundary ``flat'' locally around the data point $\bar{\inst}_i$ on the boundary, which is closest to $\inst_i$. Formally, by analyzing the limiting case $\varepsilon \rightarrow \gamma_i$, we can assume that the decision boundary induced by $\hyp$ resembles the tangent hyperplane $\hat{\mathcal{H}}_{\params}$ in a small-enough neighborhood of $\bar{\inst}_i$.

\begin{theorem}
\label{thm: epsilon-vcp non-linear model}
[Local $\varepsilon$-VCP for Non-Linear Models Under Generic Perturbation Distribution $\varphi$] Let $\model$ be a classifier and $\inst_i \in \X$ a generic input. Let $\hat{\mathcal{H}}_{\params}$ be the hyperplane tangent that locally approximates the decision boundary $\mathcal{H}_{\params}$ induced by $\hyp$ in $\bar{\inst}_i$, which is the closest data point to $\inst_i$ on the boundary. , i.e., $\bar{\inst}_i \in \mathcal{H}_{\params}$. Then:
\begin{equation}
    p_i^{\varepsilon}
= \displaystyle
  \frac{2^{\frac{n-1}{2}}}{n(n+1)B\bigl(\frac{n+1}{2}, \frac{1}{2}\bigl)\gamma_i^\frac{n-1}{2}}\delta^{\frac{n-1}{2}} + \mathcal{O}\bigl(\delta^{\frac{n+1}{2}}\bigl)\,,
\end{equation}
where $\delta = \varepsilon - \gamma_i$ and $B(\cdot,\cdot)$ is the Euler Beta function.
\begin{proof}[Proof Sketch]
    In a small-enough neighborhood of $\bar{\inst}_i$ we can approximate the decision boundary with the the tangent hyperplane $\hat{\mathcal{H}}_{\params}$. Furthermore, we can also assume the probability density $q_{\varphi}$ to remain constant within this neighborhood. Consequently Theorem \eqref{thm: epsilon-vcp linear model uniform distribution} holds. Noting that in the limiting case as $\varepsilon \rightarrow \gamma_i$, we have $\varepsilon - \gamma_i \rightarrow 0$, a Taylor expansion of the expression around zero yields the desired result. The detailed proof is provided in Appendix~\ref{app:theorem-non-linear}.
\end{proof}
\end{theorem}

\subsection{$\varepsilon$-Valid Counterfactual Probability as a Proxy of Model Generalizabilty}
\label{subsec:measure}

\begin{figure}
 \centering
 \begin{subfigure}[htb!]{.2\linewidth}
 \includegraphics[width=\linewidth]{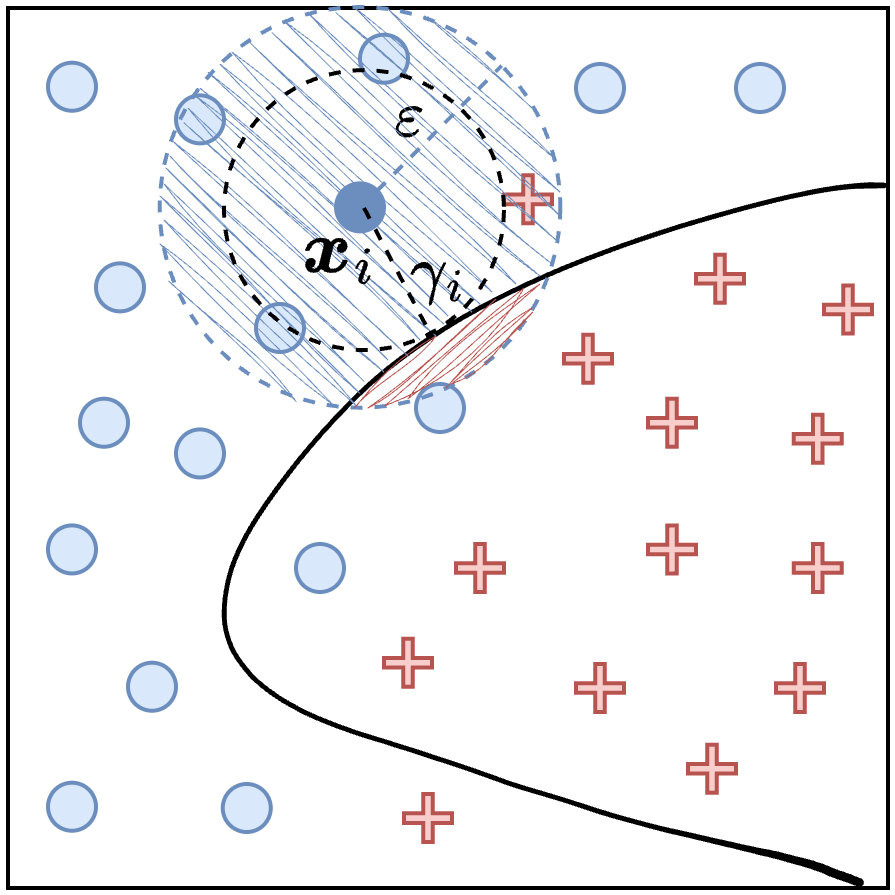}
 \caption{Low $\varepsilon$-valid counterfactual probability}
 \end{subfigure}
 \qquad
 \begin{subfigure}[h]{.2\linewidth}
 \includegraphics[width=\linewidth]{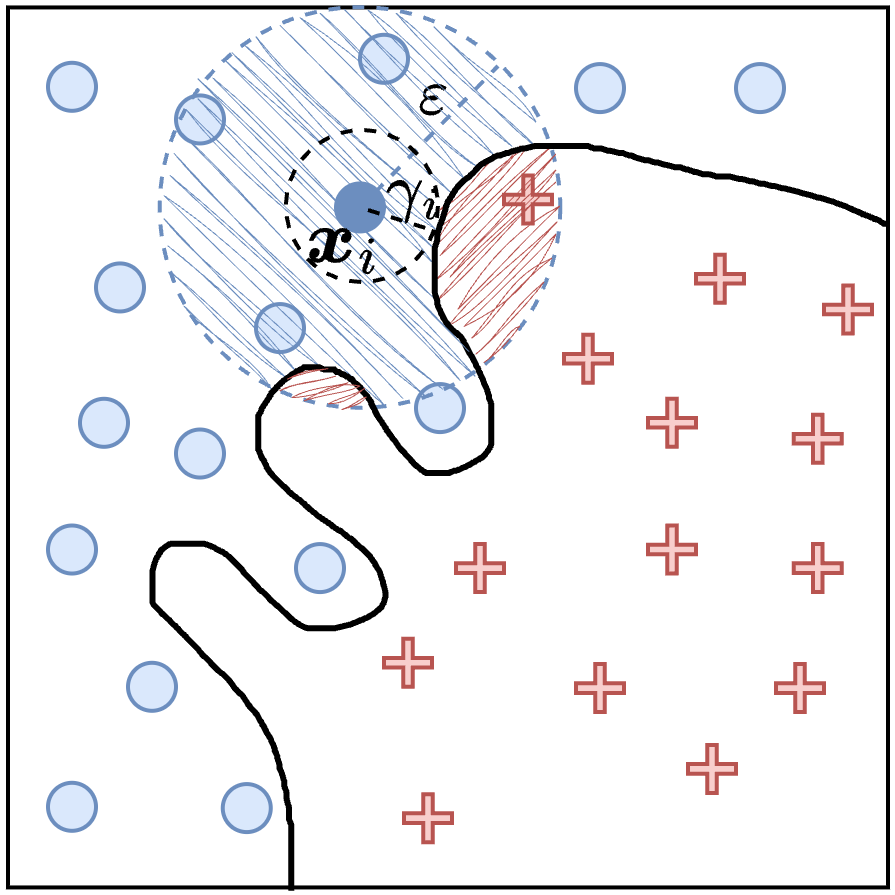}
 \caption{High $\varepsilon$-valid counterfactual probability}
 \end{subfigure}%
 \caption{The $\varepsilon$-\textit{valid counterfactual probability} for a sample $\inst \in \R^2$ can be estimated as the ratio of the area of the circle centered in $\inst$ with radius $\varepsilon$ that falls behind the decision boundary (in red).}
 \label{fig:cf-probability}
 \end{figure}

The theoretical results above capture the geometric intuition that a small margin between a data point and the decision boundary induced by a classifier leads to a higher $\varepsilon$-VCP, i.e., higher probability of finding counterfactuals. 
Indeed, as margin $\gamma_i$ gets smaller, the $\varepsilon$-VCP increases.
In other words, given a fixed $\inst_i$ and two different models having different decision boundaries (one simpler than the other), it will generally be much easier to find an $\varepsilon$-valid counterfactual for $\inst_i$ if the volume of the region where predictions are consistent with $\inst_i$, thereby $\gamma_i$, is smaller. 
This intuition is depicted in Figure~\ref{fig:cf-probability}, which illustrates how the probability of finding an $\varepsilon$-valid counterfactual for a $2$-dimensional data point may increase with model complexity, thus the risk of overfitting.

We can extend the reasoning above to the $m$ training data points and compute the average $\varepsilon$-VCP as :
\begin{equation}
\label{eq:avg-eps-vcp}
\bar{p}^{\varepsilon} = \frac{1}{m} \sum_{i=1}^m p_i^{\varepsilon}.
\end{equation}
Let us substitute the expression for exact $p_i^{\varepsilon}$ obtained in \eqref{eq:evcp-lin-uniform}, and observe that this is a function of the variable $\gamma_i$, namely:
\[
p_i^{\varepsilon} = g(\gamma_i) = \frac{1}{2}  \frac{\varepsilon^n}{\varepsilon^n - \gamma_i^n} I\Big(1-\Big(\frac{\gamma_i}{\varepsilon}\Big)^2;\frac{n+1}{2},\frac{1}{2}\Big).
\]
Specifically, $g(\gamma_i)$ is a composition of non-linear, convex, and monotonic functions for $\gamma_i \in (0, \varepsilon)$. Therefore, we can apply Jensen's inequality and obtain the following lower bound for $\bar{p}^{\varepsilon}$:
\begin{equation}
\label{eq:avg-eps-cfp}
\bar{p}^{\varepsilon} = \frac{1}{m} \sum_{i=1}^m p_i^{\varepsilon} = \frac{1}{m}\sum_{i=1}^m g(\gamma_i) \geq g \Bigg( \frac{1}{m} \sum_{i=1}^m \gamma_i \Bigg) = g(\bar{\gamma}),
\end{equation}
where $\bar{\gamma}$ denotes the average geometric margin computed across the $m$ training points. 
In Appendix~\ref{app:avg-margin}, we demonstrate that the function $g(\bar{\gamma})$ is monotonically decreasing on $\bar{\gamma}\in (0,\varepsilon)$. This confirms that as the average margin gets smaller (i.e., closer to $0$), the average $\varepsilon$-VCP gets higher.

Relevant literature on margin theory has shown that overfitting generally results in reduced margins around training samples on average (e.g., see \cite{mason1998neurips,neyshabur2017neurips,wu2019arxiv,shamir2021dimpled}).
Therefore, the average $\varepsilon$-VCP can indeed serve as a quantitative indicator of model overfitting.

%% file: experiments.tex

\section{Empirical Assessment}
\label{sec:experiments}
In this section, we empirically validate our key findings discussed in the theoretical analysis above. 

\subsection{Experimental Setup}
\label{subsec:setup}
\textit{\textbf{Estimating $\varepsilon$-VCP in Practice.}} For a fixed value of $\varepsilon$, we estimate each $p_i^{\varepsilon}$ as the fraction of the $n$-ball $\ball(\inst_i, \varepsilon)$ that crosses the decision boundary. 
To achieve this, we apply standard Monte Carlo integration, a well-established technique that uses random sampling to numerically compute definite integrals, according to \eqref{eq:evcp}. 
Specifically, we take $1,000$ random samples for each training point $\inst_i$.
Note that this estimate assumes that data points are uniformly distributed around $\inst_i$, which resembles the case when the probability distribution $\varphi$ is uniform.
For generic $\varphi$, more accurate estimates could be obtained by sampling the $\varepsilon$-neighborhood from the manifold using advanced methods (e.g., see \citet{chen2022cikm}), though this is outside the primary scope of this work and will be explored in future research. 
It is also worth mentioning that $\inst_i$ can be viewed as an embedding resulting from an autoencoding process, mapping each raw training point into a dense, lower-dimensional latent manifold within the original high-dimensional space.
\\
\textit{\textbf{Choosing $\varepsilon$.}} The choice of $\varepsilon$ depends on the underlying dataset. A good practice is to empirically determine this value to ensure a meaningful assessment of counterfactual validity while maintaining consistency with the dataset structure. Specifically, since $\varepsilon$ represents the maximum norm of the perturbation vector applied to each sample, it serves as a threshold that defines the allowable search space for counterfactuals.  
A value that is too small would overly restrict counterfactual perturbations, limiting the ability to assess meaningful changes in predictions. Conversely, a value that is too large could result in unrealistic perturbations that deviate from the underlying data distribution.  
Further discussion on this topic can be found in Appendix~\ref{app:exp-details}.
\\
\textit{\textbf{Datasets, Models, and Hardware Specifications.}} We considered two datasets: \textit{Water Potability} \citep{kadiwal2020waterpotability} and \textit{Air Quality} \citep{air_quality_360}. Furthermore, we employed a linear classifier (logistic regression) and a non-linear multi-layer perceptron (MLP) with two and five hidden layers. The logistic regression model was trained via stochastic gradient descent, while the MLP was optimized with Adam. Both models were trained for $6,000$ epochs with a learning rate of $\eta=0.001$. 
All experiments were run on a GPU NVIDIA GeForce RTX 4090 and an AMD Ryzen 9 7900 12-Core CPU. 
Full experimental details are presented in Appendix~\ref{app:exp-details}.

\subsection{Key Results}
\label{subsec:results}
\begin{figure}[ht]
    \centering
        \begin{subfigure}[t]{0.4\linewidth}
        \centering
        \includegraphics[width=\linewidth, keepaspectratio]{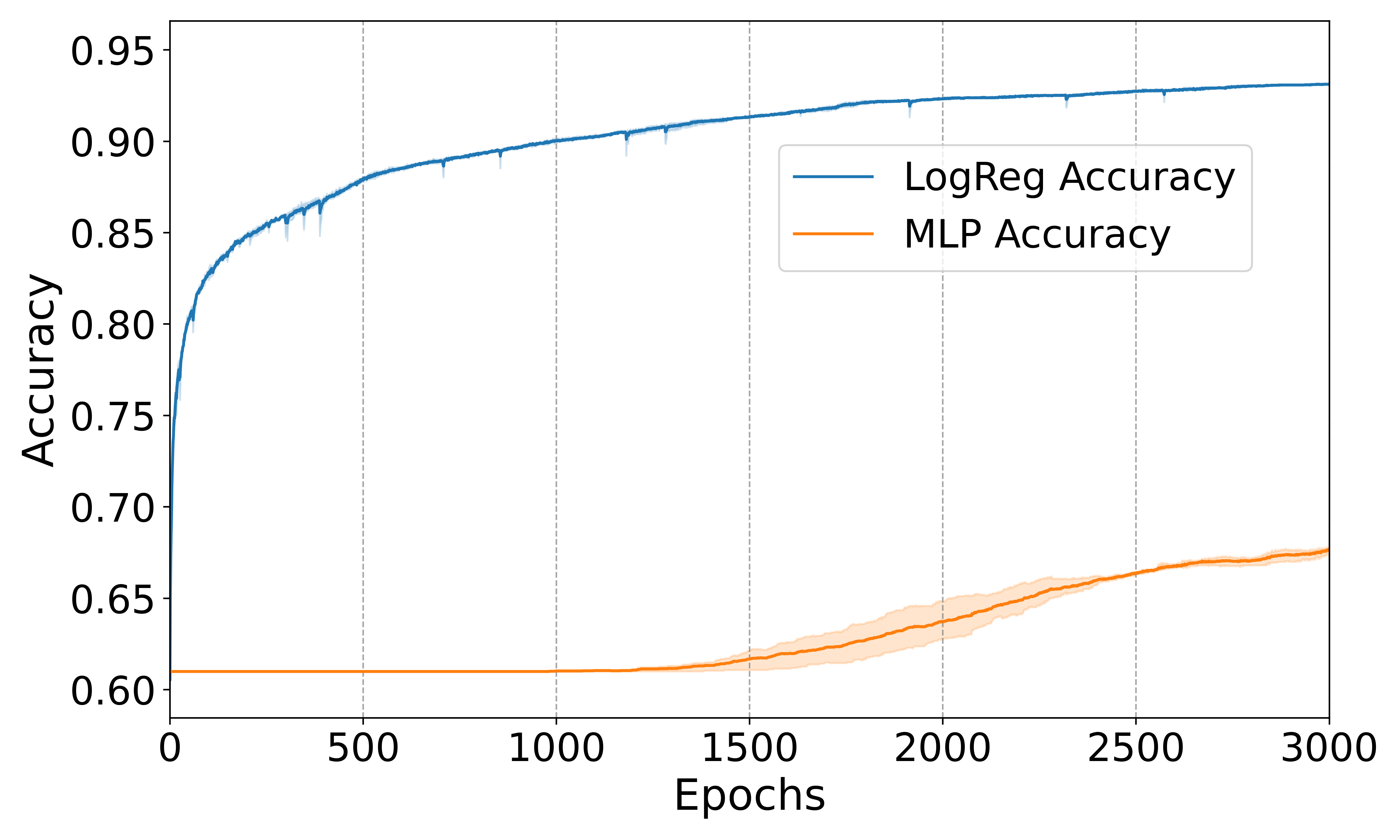}
        \caption{Training accuracy vs.\ training epochs.}
        \label{fig: logreg_water_normal_plot_accuracy}
    \end{subfigure}
    \begin{subfigure}[t]{0.4\linewidth}
        \centering
        \includegraphics[width=\linewidth, keepaspectratio]{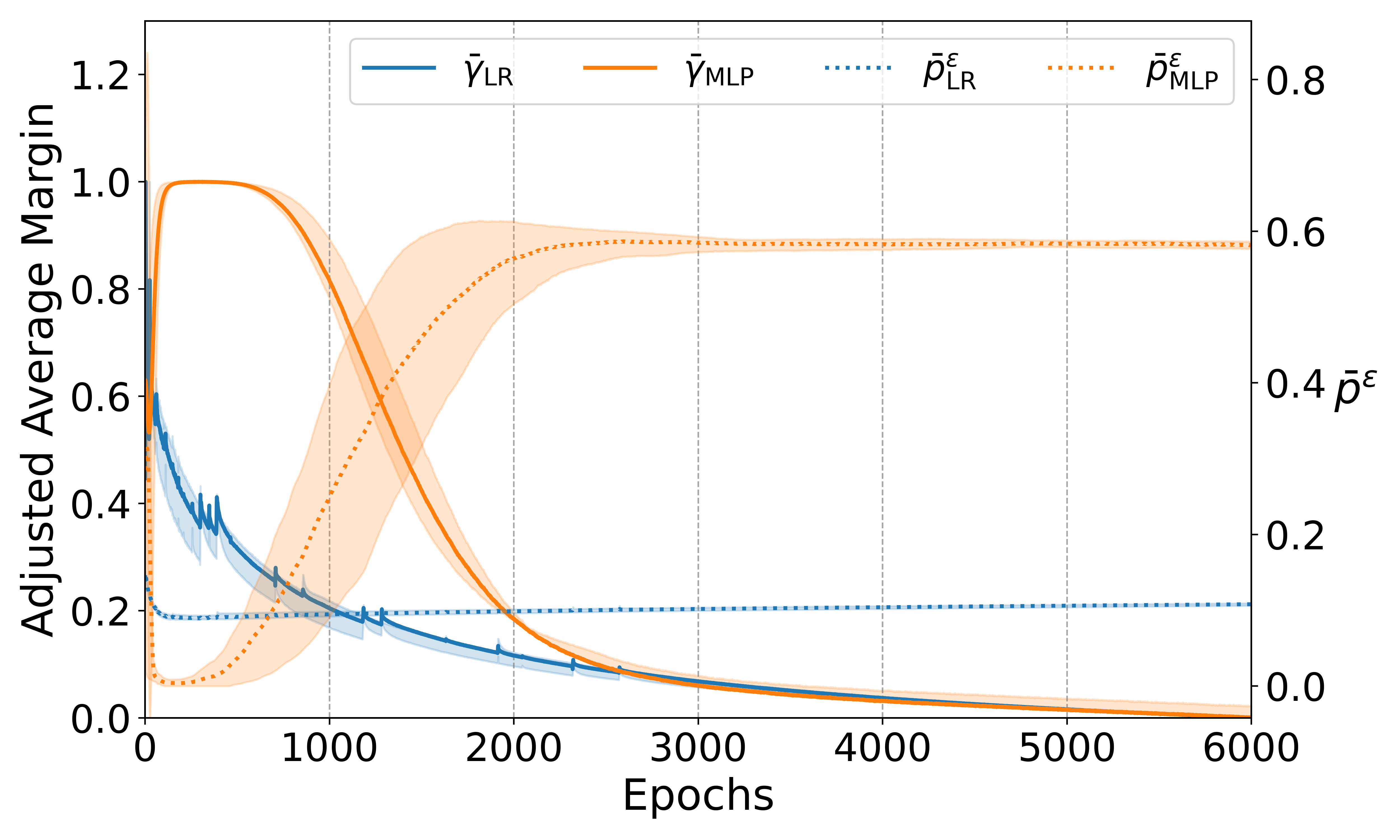}
        \caption{$\bar{p}^{\varepsilon}$ and $\bar{\gamma}$ vs.\ training epochs.}
        \label{fig: logreg_water_normal_plot_gamma_evcp}
    \end{subfigure}
    \hfill
    \caption{Evolution of training accuracy (a), $\bar{p}^{\varepsilon}$ and $\bar{\gamma}$ (b) for logistic regression and MLP.}
    \label{fig: logreg_water_normal_plot}
\end{figure}

Firstly, we show that the average $\varepsilon$-VCP increases as the average margin decreases during the course of training.
To this end, we trained a logistic regression and a two-layer MLP on the \textit{Water Potability} dataset. As expected, training accuracy steadily improves (see Figure~\ref{fig: logreg_water_normal_plot_accuracy}). At the same time, we observe a decrease in the average margin and a corresponding increase in the average $\varepsilon$-VCP.
Notice that, in the case of logistic regression, the geometric margin ($\gamma_i$) can be computed exactly for each training instance ($\inst_i$), whereas for MLP we estimate it as done by~\cite{wu2019arxiv} (see Theorem 3).
This empirical observation aligns with the theoretical relationship described in \eqref{eq:avg-eps-cfp}.

\begin{figure}[ht]
    \centering
    \begin{subfigure}[t]{0.48\linewidth}
        \centering
        \includegraphics[width=\linewidth]{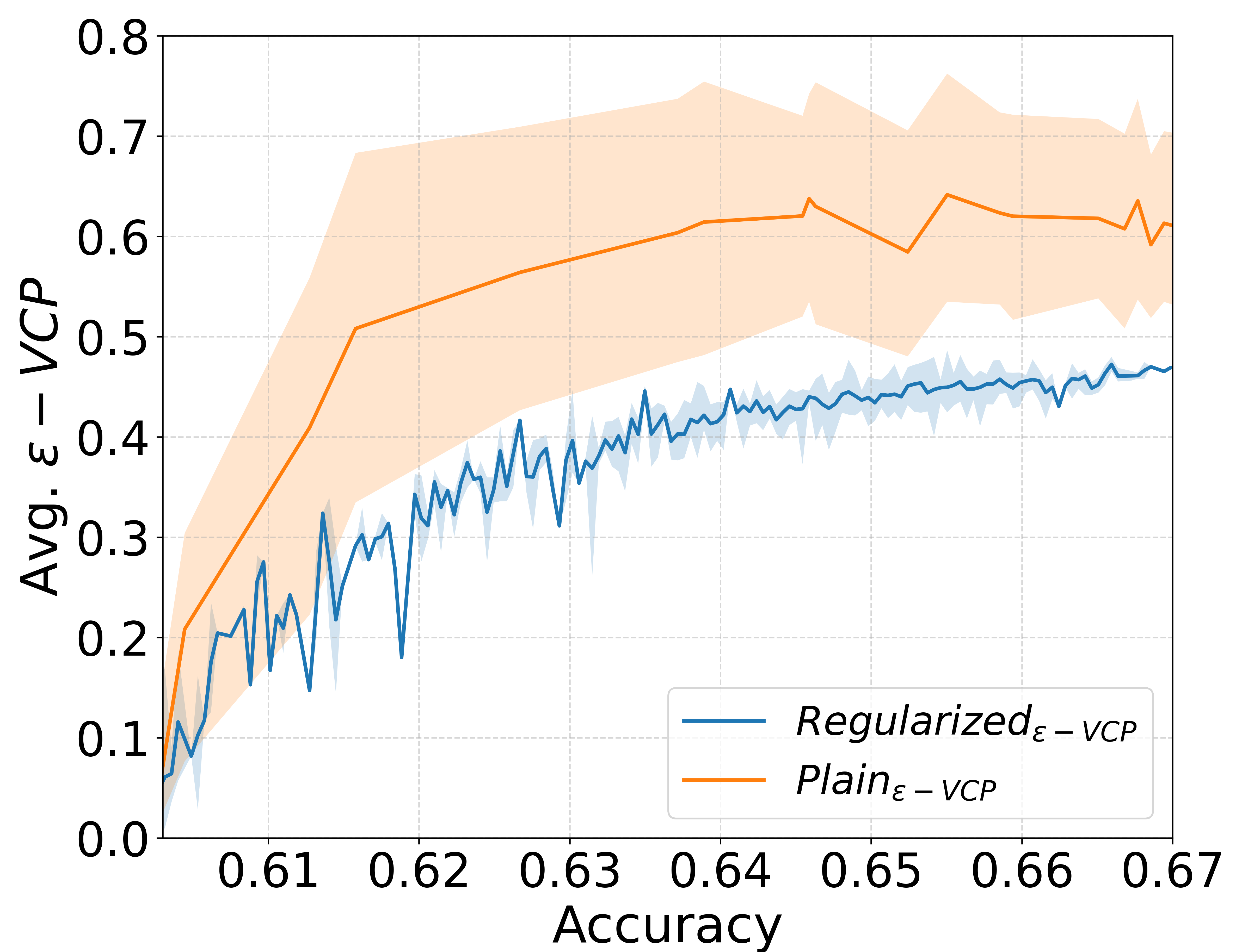}
        \caption{Average $\varepsilon$-VCP vs.\ training accuracy. \emph{Plain}$_{\varepsilon\text{-VCP}}$ is vanilla MLP; \emph{Regularized}$_{\varepsilon\text{-VCP}}$ uses $0.5$ dropout rate.}
        \label{fig:evcp-water}
    \end{subfigure}
    \hfill
    \begin{subfigure}[t]{0.48\linewidth}
        \centering
        \includegraphics[width=\linewidth, height=5.04cm]{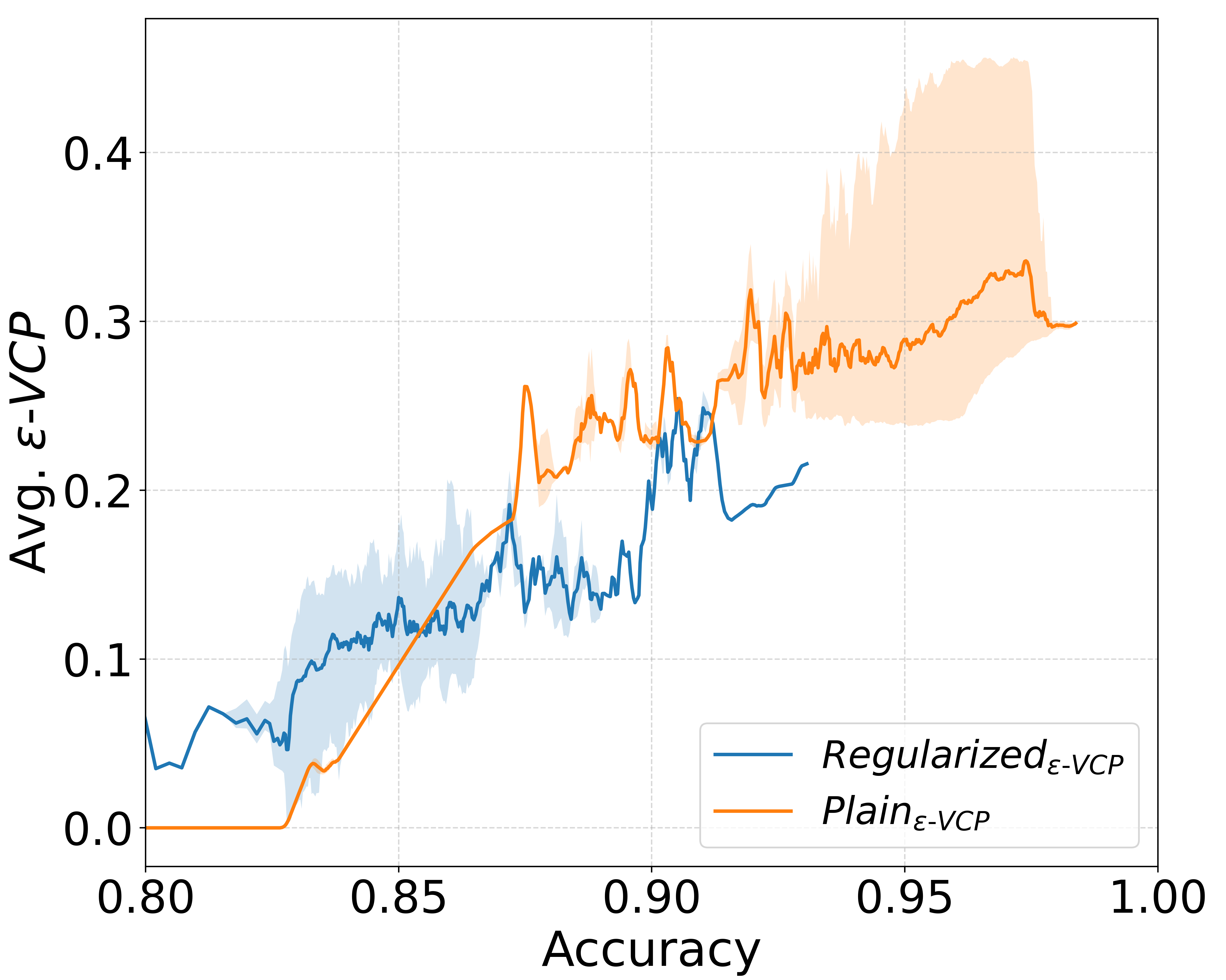}
        \caption{Average $\varepsilon$-VCP vs.\ training accuracy. \emph{Plain}$_{\varepsilon\text{-VCP}}$ is vanilla MLP; \emph{Regularized}$_{\varepsilon\text{-VCP}}$ uses $0.5$ dropout rate.}
        \label{fig:evcp-air}
    \end{subfigure}
    \caption{Comparison of the $\varepsilon$-VCP across two different datasets: (a) \textit{Water} and (b) \textit{Air Quality}.}
    \label{fig:evcp}
\end{figure}

To further investigate the impact of overfitting on the average $\varepsilon$-VCP, we trained two five-layer MLPs on the \textit{Water Potability} dataset and two on the \textit{Air Quality} dataset. 
For each pair of models, only one applied dropout regularization with a dropout rate of $0.5$. 
Figure~\ref{fig:evcp} illustrates the evolution of the average $\varepsilon$-VCP as a function of the models' training accuracy.
From this plot, three key insights emerge.
First, the average $\varepsilon$-VCP increases alongside training accuracy for both models, confirming our hypothesis that the likelihood of finding valid counterfactuals rises as models tend to overfit.
Second, unregularized MLPs consistently yield higher $\varepsilon$-VCP values compared to their regularized counterparts, indicating that their more complex decision boundaries facilitate the generation of valid counterfactuals.
Third, while regularized MLPs exhibit a less steep increase in $\varepsilon$-VCP, the effect is only mitigated. This suggests that dropout smooths the decision boundary to some extent, but may not be enough to fully counteract overfitting, leaving room for more targeted regularization strategies.

Note that, in these experiments, we track the trend of the average $\varepsilon$-VCP across training epochs, rather than computing it only at the end of training. Therefore, these empirical findings reinforce the existence of a trade-off between model generalizability and counterfactual explainability, as theoretically characterized in the previous section. 

%% file: limitations.tex
\section{Limitations and Future Work}
\label{sec:limitations}
Although our findings establish a compelling connection between generalization and counterfactual explainability, some limitations remain, which open avenues for future work.
\\
First, our theoretical analysis relies on the geometric margin $\gamma_i$, which can be computed exactly only for linear classifiers. For more complex models, a potential direction is to estimate the margin using a $K$-Lipschitz assumption on $\hyp$, using techniques similar to those proposed by \cite{hein2017formal} and \cite{tsuzuku2018lipschitz} for adversarial robustness.
\\
Second, we assume uniform perturbations to compute $\varepsilon$-VCP, which may not reflect real-world data geometry. In practice, data often lies on low-dimensional manifolds embedded in high-dimensional spaces, and meaningful perturbations should ideally respect this geometry. Future work could incorporate manifold-based or data-driven perturbations (e.g., \cite{stutz2019disentangling}) to compute more realistic counterfactuals and obtain sharper insights into the generalization–explainability trade-off.
\\
Finally, in future work, we plan to design a regularization term based on $\varepsilon$-VCP and incorporate it into standard training objectives to mitigate overfitting while preserving explainability.


%% file: conclusion.tex
\section{Conclusion}
\label{sec:conclusion}
In this work, we investigated the connection between model generalization and counterfactual explainability in supervised learning. We showed that the average probability of generating counterfactual examples over the training set, quantified by the average $\varepsilon$-VCP, increases with model overfitting. This establishes a rigorous link between poor generalization and the ease of counterfactual generation, revealing an inherent trade-off between generalization and counterfactual explainability.
Our theoretical insights were supported by empirical evidence, suggesting that $\varepsilon$-VCP can serve as a practical and quantitative proxy for measuring model overfitting.

%% file: supplementary.tex
\section{Technical Appendices and Supplementary Material}

\subsection{Full Proof of Theorem 4.3}
\label{app:theorem-non-linear}

\noindent
Let $\model$ be a classifier and $\inst_i \in \X$ a generic input. Let $\hat{\mathcal{H}}_{\params}$ be the hyperplane tangent that locally approximates the decision boundary $\mathcal{H}_{\params}$ induced by $\hyp$ in $\bar{\inst}_i$, which is the closest data point to $\inst_i$ on the boundary. , i.e., $\bar{\inst}_i \in \mathcal{H}_{\params}$. Then:
\begin{equation*}
    p_i^{\varepsilon}
= \displaystyle
  \frac{2^{\frac{n-1}{2}}}{n(n+1)B\bigl(\frac{n+1}{2}, \frac{1}{2}\bigl)\gamma_i^\frac{n-1}{2}}\delta^{\frac{n-1}{2}} + \mathcal{O}\bigl(\delta^{\frac{n+1}{2}}\bigl)\,,
\end{equation*}
where $\delta = \varepsilon - \gamma_i$ and $B(\cdot,\cdot)$ is the Euler Beta function.
\begin{proof}
Let \( \delta = \varepsilon - \gamma_i \), so that \( \varepsilon = \gamma_i + \delta \). We analyze the behavior of \( p_i^{\varepsilon} \) as \( \delta \to 0 \).

The argument of the regularized incomplete beta function is
\[
x = \frac{(\varepsilon - \gamma_i)(\varepsilon + \gamma_i)}{\varepsilon^2} = \frac{\delta (2\gamma_i + \delta)}{(\gamma_i + \delta)^2}.
\]
Expanding the denominator:
\[
(\gamma_i + \delta)^2 = \gamma_i^2 + 2\gamma_i \delta + \delta^2.
\]
Therefore,
\[
x = \frac{2\gamma_i \delta + \delta^2}{\gamma_i^2 + 2\gamma_i \delta + \delta^2} = \frac{2\gamma_i \delta}{\gamma_i^2} + o(\delta) = \frac{2\delta}{\gamma_i} + \mathcal{O}(\delta).
\]

For small \( x \), the regularized incomplete beta function satisfies
\[
I(x;a,b) = \frac{x^a}{a B(a,b)}\bigr[ 1+ \mathcal{O}(x)\bigr] \quad \text{as } x \to 0.
\]
Applying this with \( a = \tfrac{n+1}{2} \) and \( b = \tfrac{1}{2} \), we obtain
\begin{equation*}
    \begin{aligned}
        I\left(\frac{2\delta}{\gamma_i}; \tfrac{n+1}{2}, \tfrac{1}{2} \right) &= \frac{1}{\tfrac{n+1}{2} B\left( \tfrac{n+1}{2}, \tfrac{1}{2} \right)} \left( \frac{2\delta}{\gamma_i} \right)^{\tfrac{n+1}{2}}\bigr[ 1+ \mathcal{O}(\delta)\bigr]\\
        &=\frac{1}{\tfrac{n+1}{2} B\left( \tfrac{n+1}{2}, \tfrac{1}{2} \right)} \left( \frac{2\delta}{\gamma_i} \right)^{\tfrac{n+1}{2}}+ \mathcal{O}\bigl(\delta^{\tfrac{n+3}{2}}\bigl)
    \end{aligned}
\end{equation*}

About \( \varepsilon^n \) and \( \varepsilon^n - \gamma_i^n \), using the binomial expansion we have:
\[
\varepsilon^n = (\gamma_i + \delta)^n = \gamma_i^n + n \gamma_i^{n-1} \delta + \mathcal{O}(\delta),
\]
so
\[
\varepsilon^n - \gamma_i^n = n \gamma_i^{n-1} \delta + \mathcal{O}(\delta).
\]

Combining the expansion in equation \eqref{eq:evcp-lin-uniform} takes us to
\begin{align*}
p_i^{\varepsilon} &= \frac{\tfrac{1}{2}  (\gamma_i^n + n \gamma_i^{n-1} \delta + \mathcal{O}(\delta)) \cdot \biggr[\frac{1}{\tfrac{n+1}{2} B\left( \tfrac{n+1}{2}, \tfrac{1}{2} \right)} \left( \frac{2\delta}{\gamma_i} \right)^{\tfrac{n+1}{2}}+\mathcal{O}\bigl(\delta^{\tfrac{n+3}{2}}\bigl)\biggr]}{ n \gamma_i^{n-1} \delta+\mathcal{O}(\delta)} \\
&= \frac{\tfrac{1}{2}  \gamma_i^n  \cdot \frac{1}{\tfrac{n+1}{2} B\left( \tfrac{n+1}{2}, \tfrac{1}{2} \right)} \left( \frac{2\delta}{\gamma_i} \right)^{\tfrac{n+1}{2}}+\mathcal{O}\bigl(\delta^{\tfrac{n+3}{2}}\bigl)}{ n \gamma_i^{n-1} \delta+\mathcal{O}(\delta)}.\\
&= \frac{\tfrac{1}{2}  \gamma_i^{n-(\frac{n+1}{2})}  \cdot \frac{2^{\frac{n+1}{2}}}{\tfrac{n+1}{2} B\left( \tfrac{n+1}{2}, \tfrac{1}{2} \right)} \left( \delta \right)^{\tfrac{n+1}{2}}+\mathcal{O}\bigl(\delta^{\tfrac{n+3}{2}}\bigl)}{ n \gamma_i^{n-1} \delta+\mathcal{O}(\delta)}.\\
&= \frac{\Biggr[\tfrac{1}{2}  \gamma_i^{n-(\frac{n+1}{2})}  \cdot \frac{2^{\frac{n+1}{2}}}{\tfrac{n+1}{2} B\left( \tfrac{n+1}{2}, \tfrac{1}{2} \right)} \left( \delta \right)^{\tfrac{n+1}{2}}\Biggr]\cdot \Bigr(1+\mathcal{O}\bigl(\delta\bigl)\Bigr)}{ n \gamma_i^{n-1} \delta \cdot \bigr(1 +\mathcal{O}(1)\bigr)}.\\
&= \frac{2^{\frac{n-1}{2}}}{n(n+1)B\bigl(\frac{n+1}{2}, \frac{1}{2}\bigl)\gamma_i^\frac{n-1}{2}}\delta^{\frac{n-1}{2}} + \mathcal{O}\bigl(\delta^{\frac{n+1}{2}}\bigl).
\end{align*}

\end{proof}

\subsection{Average $\varepsilon$-VCP is Monotonically Decreasing with the Average Margin}
\label{app:avg-margin}

   \noindent
    The function 
    \begin{equation*}
        g(\bar{\gamma})
  =  \frac12\,\frac{\varepsilon^n}{\varepsilon^n - \bar{\gamma}^n}\,
    I\Bigl(1 - \bigl(\tfrac{\bar{\gamma}}{\varepsilon}\bigr)^2;\tfrac{n+1}{2},\tfrac12\Bigr),
    \end{equation*}
    is monotonically decreasining in $\quad 0 < \bar{\gamma} < \varepsilon$.

    \begin{proof}
Let us define two components:
\begin{align*}
    A(\bar{\gamma}) &= \frac{\varepsilon^n}{\varepsilon^n - \bar{\gamma}^n}, \\
    B(\bar{\gamma}) &= I_{1 - \bar{\gamma}/\varepsilon}\left( \frac{n+1}{2}, \frac{1}{2} \right).
\end{align*}
Then:
\[
    g(\bar{\gamma}) = \frac{1}{2} A(\bar{\gamma}) B(\bar{\gamma})
\]
Differentiating using the product rule:
\[
    g'(\bar{\gamma}) = \frac{1}{2} \left[ A'(\bar{\gamma}) B(\bar{\gamma}) + A(\bar{\gamma}) B'(\bar{\gamma}) \right].
\]

We have that:
\[
    A'(\bar{\gamma}) = \frac{d}{d\bar{\gamma}} \left( \frac{\varepsilon^n}{\varepsilon^n - \bar{\gamma}^n} \right) = \frac{n \varepsilon^n \bar{\gamma}^{n-1}}{(\varepsilon^n - \bar{\gamma}^n)^2} > 0
\]

and, set \( x = 1 - \bar{\gamma}/\varepsilon \), so that:
\[
    B(\bar{\gamma}) = I_{x}(a, b), \quad a = \frac{n+1}{2},\; b = \frac{1}{2}.
\]
using the chain rule and the known derivative:
\[
    \frac{d}{d\bar{\gamma}} B(\bar{\gamma}) = -\frac{2}{\varepsilon} \cdot \frac{x^{a-1}(1 - x)^{b-1}}{\mathrm{B}(a, b)} < 0,
\]
because all terms are positive and the negative sign comes from \( dx/d\bar{\gamma} = -1/\varepsilon \).

Since
\[
    g'(\bar{\gamma}) = \frac{1}{2} \left[ A'(\bar{\gamma}) B(\bar{\gamma}) + A(\bar{\gamma}) B'(\bar{\gamma}) \right].
\]
we are asking if  AB', which is negative, dominates over the positive term A'B.

\noindent
Let's provide a qualitative argument by studying the  behavior near edges of the domain.

\noindent
For $\bar{\gamma} \rightarrow 0$ we have:
\begin{equation*}
    \begin{aligned}
       A(\bar{\gamma})&\rightarrow1\,,\\
       A'(\bar{\gamma})&\rightarrow0\,,\\
       B(\bar{\gamma})&\rightarrow1\,,\\
       B'(\bar{\gamma})&\rightarrow \text{finite negative number}.
    \end{aligned}
\end{equation*}
hence, 
\[
    g'(\bar{\gamma}) \approx \frac{1}{2} \left[ 0\cdot 1  + 1\cdot (\text{negative}) \right] < 0.
\]
For $\bar{\gamma} \rightarrow \varepsilon$ we have:
\begin{equation*}
    \begin{aligned}
       A(\bar{\gamma})&\rightarrow +\infty\,,\\
       A'(\bar{\gamma})&\rightarrow +\infty\,,\\
       B(\bar{\gamma})&\rightarrow0\,,\\
       B'(\bar{\gamma})&\rightarrow \text{finite negative number}.
    \end{aligned}
\end{equation*}
about the products, for $n>3$ we have:
\begin{equation*}
    \begin{aligned}
       A'B&\rightarrow0\,,\\
       AB'&\rightarrow +\infty\cdot(\text{finite negative number}) = -\infty.
    \end{aligned}
\end{equation*}
with $A'B\rightarrow0$ being computed using the expansion of the regularized incomplete beta function as we have done in proof of theorem \eqref{thm: epsilon-vcp non-linear model}.

Hence, again, we have 
\[
    g'(\bar{\gamma}) \approx \frac{1}{2} \left[ 0  -\infty \right] < 0.
\]

Inside the domain, the negative term in $g'(\bar{\gamma})$ continues to dominate the positive term, as it does at the boundary. As a consequence, $g(\bar{\gamma})$ is monotonically decreasing over the entire interval $(0, \varepsilon)$. To further support this claim, we provide empirical evidence in Figure~\ref{fig: monotonicity_bound_g}, demonstrating the monotonic behavior of $g(\bar{\gamma})$ for various values of $n$ and $\varepsilon$.

\begin{figure}[ht]
    \centering
        \begin{subfigure}[t]{0.4\linewidth}
        \centering
        \includegraphics[width=\linewidth, keepaspectratio]{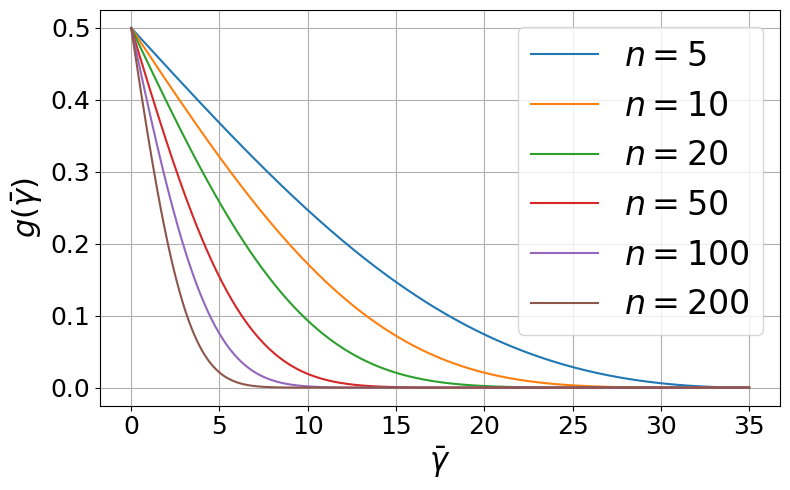}
        \caption{The impact of $n$ on $g(\bar{\gamma})$ for $\varepsilon = 35$.}
        \label{fig: effect_n_bound}
    \end{subfigure}
    \begin{subfigure}[t]{0.4\linewidth}
        \centering
        \includegraphics[width=\linewidth, keepaspectratio]{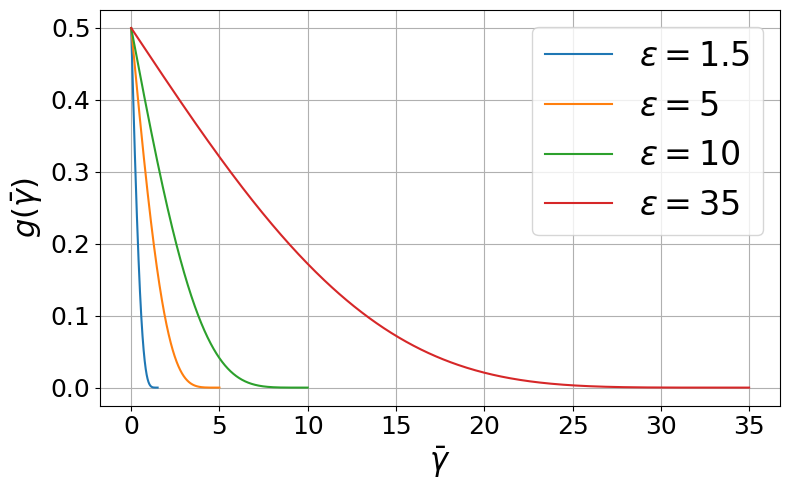}
        \caption{The impact of $\varepsilon$ on $g(\bar{\gamma})$ for $n = 10$.}
        \label{fig: effect_epsilon_bound}
    \end{subfigure}
    \hfill
    \caption{Monotonicity of $g(\bar{\gamma}).$}
    \label{fig: monotonicity_bound_g}
\end{figure}

    \end{proof}

\subsection{Experimental Details}
Before training the model, we apply standard preprocessing by normalizing each input feature to have zero mean and unit variance.
To intentionally induce overfitting in our logistic regression model, we perform a polynomial basis expansion of degree $d = 6$. This transformation generates all polynomial combinations of the original features whose total degree does not exceed $d$, including both higher-order terms of individual features and multi-feature interaction terms. The resulting high-dimensional feature space significantly increases the number of parameters in the model, making it more susceptible to overfitting.

As a consequence of this expansion, we rescale the maximum perturbation value $\varepsilon$ to account for the increased dimensionality, ensuring the perturbation magnitude remains consistent across different feature spaces.
Specifically, given the original perturbation budget $\varepsilon_{\text{ori}}$ corresponding to $n$ features, we adjust it to $\varepsilon_{\text{aug}}$ for the expanded feature space with $n'$ features using the following rule:
\[
\varepsilon_{\text{aug}} = \varepsilon_{\text{ori}}\sqrt{\frac{n'}{n}}.
\]
In this way, we can compute the value for $\varepsilon$ regardless of the dimensionality of the feature space. 

Table \ref{tab:exp-params} reports all the parameter settings for every experiment presented in the main paper.

\begin{table*}[ht]
\centering
\caption{Summary of experimental settings.}
\label{tab:exp-params}
\scalebox{0.85}{
\begin{tabular}{lcccc}
\toprule
\textbf{Parameter}                
& \shortstack[l]{\textbf{Exp.~LogReg}\\\textbf{(Fig.~\ref{fig: logreg_water_normal_plot_accuracy}, \ref{fig: logreg_water_normal_plot_gamma_evcp})}}
 & \shortstack[l]{\textbf{Exp.~MLP}\\\textbf{(Fig.~\ref{fig: logreg_water_normal_plot_accuracy}, \ref{fig: logreg_water_normal_plot_gamma_evcp})}} 
 & \shortstack[l]{\textbf{Exp.~Water}\\\textbf{(Fig.~\ref{fig:evcp-water})}} 
  & \shortstack[l]{\textbf{Exp.~Air Quality}\\\textbf{(Fig.~\ref{fig:evcp-water})}} \\
\midrule
Model                                 & LogReg    &    MLP                   &    MLP                  & MLP\\
Layers                                & None         &    [100, 30]             &    [100, 50, 25, 15, 5] &  [100, 50, 25, 15, 5]             \\
Dataset                               & Water        &    Water                 &    Water                & Air Quality\\
Train/Test split                    & $0.8/0.2$      &   $0.8/0.2$                &   $0.8/0.2$               & $0.8/0.2$\\
Feature standardization               & yes          & yes                      & yes                     & yes \\
Polynomial expansion degree ($d$)     &  $6$           &    None                  &   None                  &  None \\
Input feat. dim. ($n$)         &  $9$           &    $9$                     &   $9$                     & $14$\\
Augmented feat. dim. ($n'$)    &  $5,005$                 &    $9$                     &   $9$            &  $14$\\
Perturbation radius ($\varepsilon$)&   $35.00$       &    $1.500$                   &   $1.500$          &  $0.373$  \\
Regularization (type \& strength)     &   None                &     None                 &   None/Dropout ($0.5$) &   None/Dropout ($0.5$)     \\
Solver/Optimizer                    &   SGD                 &     SGD                  &   Adam              & Adam \\
Learning rate                         &  $0.001$                &     $0.001$                &   $0.001$              & $0.00001$\\
Batch size                            &   $128$                &     $128$                  &   $128$              & $128$\\
Number of epochs                      &   $6,000$                &     $6,000$                 &   $6,000$               & $6,000$\\
\bottomrule
\end{tabular}
}
\end{table*}